\documentclass[10pt,twocolumn,letterpaper]{article}

\usepackage{iccv}
\usepackage{times}
\usepackage{epsfig}
\usepackage{graphicx}
\usepackage{amsmath}
\usepackage{amssymb}

\usepackage[breaklinks=true,bookmarks=false]{hyperref}

\iccvfinalcopy 

\def\iccvPaperID{****} 

\ificcvfinal\fi

\begin{document}

\title{TBN-ViT: Temporal Bilateral Network with Vision Transformer for Video Scene Parsing}

\author{Bo Yan, Leilei Cao, Hongbin Wang\\
Ant Group\\
}

\maketitle
\ificcvfinal\thispagestyle{empty}\fi

\begin{abstract}
   Video scene parsing in the wild with diverse scenarios is a challenging and great significance
task, especially with the rapid development of automatic driving technique. The dataset Video Scene Parsing in the Wild(VSPW) contains well-trimmed long-temporal, dense annotation and high resolution clips. Based on VSPW, we design a Temporal Bilateral Network with Vision Transformer. We first design a
spatial path with convolutions to generate low level features which can preserve the spatial information. Meanwhile, a context path with vision transformer is employed to obtain sufficient context information. Furthermore, a temporal context module is designed to harness the inter-frames contextual information. Finally, the proposed method can achieve the mean intersection over union(mIoU) of 49.85\% for the VSPW2021 Challenge test dataset.

\end{abstract}

\section{Introduction}

Video scene parsing applies widely in video editing, video composition, autonomous driving, etc. The video scene parsing is primarily a task of video semantic segmentation, but it is more difficult and challenging due to the wide range of real-world scenarios and dense annotation for every pixel, that's to say we must segment the specific categories for backgrounds(e.g., road, wall, sky). VSPW~\cite{miao2021vspw} is a suitable benchmark to tackle the challenging video scene parsing task.

Semantic segmentation is a fundamental problem in computer vision. Deep learning based methods have achieved promising results for image semantic segmentation over the past few years, such as DeepLab series~\cite{7913730, DBLP:journals/corr/ChenPSA17, Chen_2018_ECCV}, OCRNet~\cite{10.1007/978-3-030-58539-6_11}, BiSeNet series~\cite
{Yu_2018_ECCV, DBLP:journals/corr/abs-2004-02147}, etc. Several methods are proposed to solve the video smantic segmentation task, TDNet~\cite{Hu_2020_CVPR},  TMANet~\cite{DBLP:journals/corr/abs-2102-08643} and TCBNet~\cite{miao2021vspw} are latest representative methods, which can capture  temporal information between frames by particular design of temporal module.

\hfill

Inspired by BiSeNet~\cite
{Yu_2018_ECCV} and Swin Transformer~\cite{DBLP:journals/corr/abs-2103-14030},  we design a bilateral network with vision transformer, spatial path with convolutions to generate low level features which can preserve the spatial information, and context path with vision transformer is employed to obtain sufficient context information. Finally, in order to effectively harness temporal information from the past frames, we use the method proposed by TCBNet~\cite{miao2021vspw} to get a better and more stable segment results.

\hfill

\section{Approach}

Our approach is developed base on the BiSeNet framework. The structure is illustrated in Figure 1. We introduce the proposed TBN-ViT network in Sec.2.1. The detail of training strategy is introduced in Sec.2.2.

\begin{figure*}[t]
	\centering
	\includegraphics[scale=0.4]{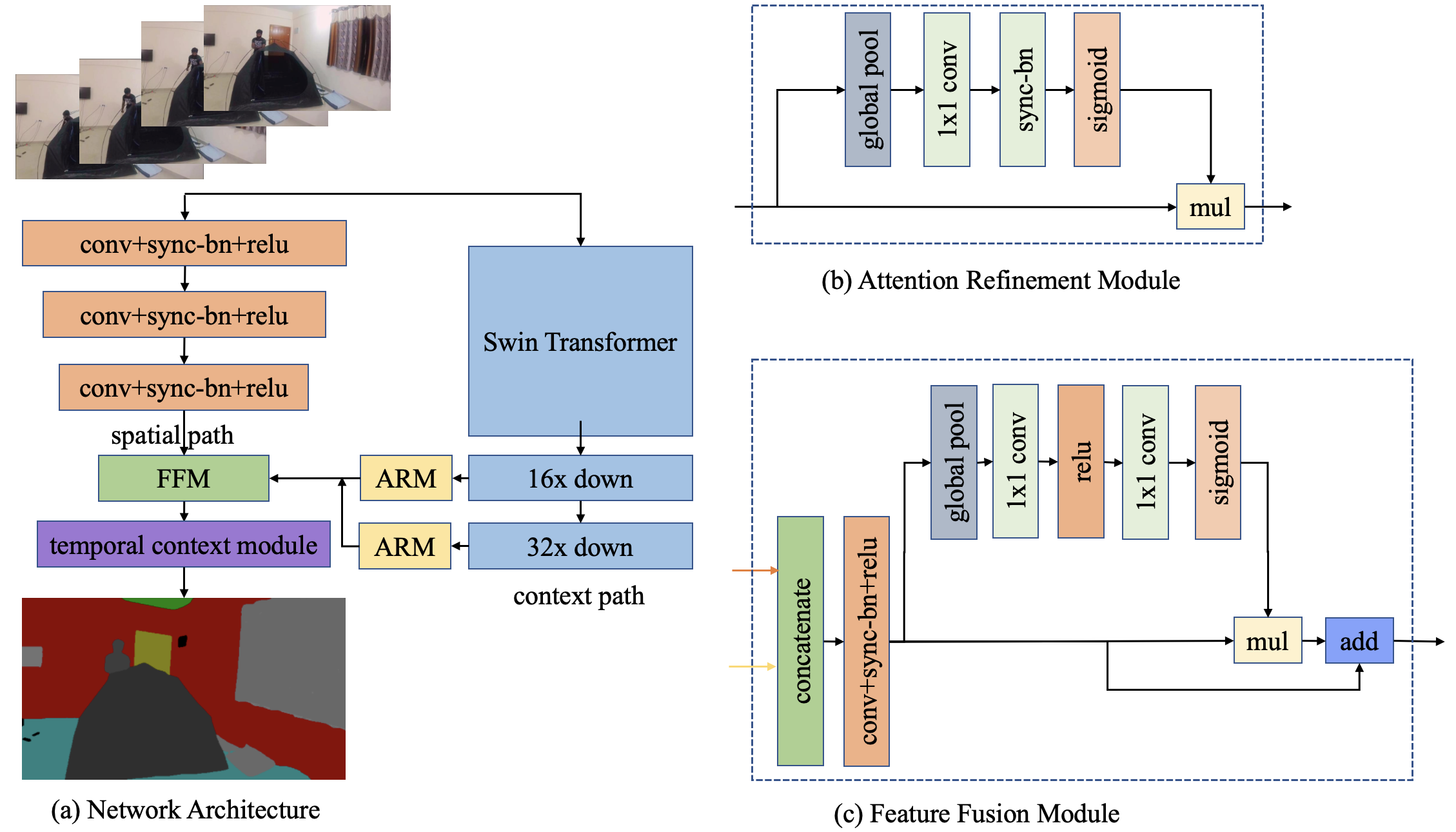}
	\caption{An overview of the Temporal Bilateral Network with Vision Transformer. \textbf{(a)} Network Architecture. \textbf{(b)} Components of the Attention Refinement Module (ARM). \textbf{(c)} Components of the Feature Fusion Module(FFM).}
	\label{pipeline}
\end{figure*}

\subsection{TBN-ViT Network}

Spatial information is important for pixel-level task such as detection and segmentation, etc. So we use three convolution layers with stride=2 to generate low level features, the original image is down sampled to 1/8 which can preserve rich spatial information.

\hfill

Transformers have made enormous strides in NLP~\cite{DBLP:journals/corr/abs-1810-04805, radford2019language}. There are quite a bit of works apply transformers to computer vision~\cite{DBLP:journals/corr/abs-1906-05909,Zhao_2020_CVPR, 10.1007/978-3-030-58452-8_13}. Especially, Swin Transformer~\cite{DBLP:journals/corr/abs-2103-14030} has reached state-of-the-art performance on the detect and segment task on COCO. Transformers can capture the non-local and relational nature of languages and images, so we consider using Swin Transformer as segmentation backbone, which can reserve rich context information. Firstly, we get our segmentation backbone by dropping the last classification layer from original Swin Transformer-Large. Then we extract the feature maps from the last and second last layer of segmentation backbone. Finally Attention Refinement Module(ARM) is applied to refine the feature maps.

\hfill

Finally, the feature maps from spatial path and context path are fused by Feature Fusion Module(FFM). Besides, in order to effectively harness temporal information from the past frames, a temporal context module is proposed. We first select last three, last six and last nine frames from current frame, and input these three frames to the whole pipeline to extract the fused feature maps, then get the temporal feature map by calculating the average of feature maps of these three frames, finally the feature map from current frame and temporal feature map are concatenated to  classify and get the final segment result.

\subsection{Training Strategy}

Due to the limitation of GPU memory(NVIDIA Tesla V100 16GB), we use 8 gpus for training but only one sample on one gpu, so Synchronized Batch Normalization(Sync-BN) is applied to get a better performance. Stochastic Gradient Descent(SGD) with Cross Entropy(CE) loss is applied in the first stage of training, SGD with Online Hard Example Mining(OHEM) CE loss is applied in the second stage of training, with this strategy the segment result can be better and more robust.

\hfill

\section{Experiments}
We train and evaluate our method on the VSPW2021 dataset.

\subsection{Training Details}
We only use the VSPW2021 train dataset for training. In the first stage of training, the pre-trained swin transformer-large model provided by Swin Transformer~\cite{DBLP:journals/corr/abs-2103-14030} is applied for context path, other layers are trained from scratch. Optimizer is SGD with learning rate(lr)=0.002 for whole network with CE loss, batch size is 8 on 8 gpus, and the input images are cropped to [479, 479]. Other data augmentation like random flip, random crop etc. are also applied. In the second stage of training, whole network are trained based on the first stage with OHEM CE loss. Optimizer is SGD with lr=0.0002 for context path and lr=0.0005 for other layers. 

\subsection{Experimental Results}
As shown in Table 1, Our proposed method finally achieves 49.85\% mIoU on the VSPW2021 Challenge test dataset.


\begin{table*}
\begin{center}
\begin{tabular}{lcc}
\hline
Methods & mIoU & Boost \\
\hline
Baseline(Spatial-Temporal PPM with ResNet-101 Backbone) & 37.46\% & -\\
Spatial-Temporal PPM with Swin Transformer-Large Backbone & 43.63\% & +6.17\% \\
Bilateral Network with Swin Transformer-Large Backbone & 45.73\% & +2.10\% \\
Temporal Bilateral Network with Swin Transformer-Large Backbone & 47.17\% & +1.44\% \\
Temporal Bilateral Network with Swin Transformer-Large Backbone + OHEM & 49.85\% & +2.68\% \\
\hline
\end{tabular}
\end{center}
\caption{Ablation study on VSPW2021 test dataset}
\end{table*}

\subsection{Ablation Study}
In this section, we elaborate how we achieve the final result by ablation study to 
explain the contribution of our proposed method. Following experiments are conducted on VSPW2021 test dataset and mIoU score is reported. The baseline is the original Spatial-Temporal PPM with ResNet-101 backbone\cite{miao2021vspw} achieves 37.46\% on VSPW2021 test dataset.

We apply Swin Transformer-Large as backbone for Spatial-Temporal PPM, which achieves better mIoU score of 43.63\%, with a significant improvement of 6.17\%. Then we demonstrate the effectiveness of our proposed Bilateral Network with Vision Transformer, which can bring an improvement of 2.10\%. Next we add temporal context module for Bilateral Network, which brings an improvement of 1.44\%. Last the OHEM CE loss is applied to finetune the model, it brings an improvement of 2.68\%. Finally we achieve 49.85\% mIoU score on the VSPW2021 test dataset.

\hfill

\section{Conclusion}

In this paper, we propose a Temporal Bilateral Network with Vision Transformer for video scene parsing, including a spatial path with convolutions, a context path with vision transformer and a temporal context module, and also a training strategy. Our approach achieves 49.85\% mIoU score on the VSPW2021 Challenge test dataset.


{\small
\bibliographystyle{ieee_fullname}
\bibliography{egbib}
}

\end{document}